\documentclass[runningheads]{llncs}
\usepackage[T1]{fontenc}
\usepackage[]{algorithm2e}
\usepackage{amsmath,amsfonts}
\usepackage{subcaption}
\usepackage{graphicx}
\begin{document}
\title{Hyperparameter Optimization for Driving Strategies Based on Reinforcement Learning}
\titlerunning{HPO for Driving Strategies based on RL}
%

\author{Nihal Acharya Adde\inst{1}\orcidID{0009-0002-0966-0825} \and Hanno Gottschalk\inst{2}\orcidID{0000-0003-2167-2028} \and
Andreas Ebert\inst{3}}
\authorrunning{Nihal Acharya Adde et al.}
%
\institute{Volkswagen Group Innovation, Volkswagen AG, Wolfsburg, Germany  \email{nihal.acharya.adde@volkswagen.de}\and
Institute of Mathematics, TU Berlin, Germany
\email{gottschalk@math.tu-berlin.de} \and
Volkswagen Group Innovation, Volkswagen AG, Wolfsburg, Germany \\
\email{andreas.ebert@volkswagen.de}}
\maketitle              

\begin{abstract}
This paper focuses on hyperparameter optimization for autonomous driving strategies based on Reinforcement Learning (RL). We provide a detailed description of training the RL agent in a simulation environment. Subsequently, we employ Efficient Global Optimization (EGO) algorithm that uses Gaussian Process (GP) fitting for hyperparameter optimization in RL. Before this optimization phase, Gaussian process interpolation is applied to fit the surrogate model, for which the hyperparameter set is generated using Latin hypercube sampling. To accelerate the evaluation, parallelization techniques are employed. Following the hyperparameter optimization procedure, a set of hyperparameters is identified, resulting in a noteworthy enhancement in overall driving performance. There is a substantial increase of 4\% when compared to existing manually tuned parameters and the hyperparameters discovered during the initialization process using Latin hypercube sampling. After the optimization, we analyze the obtained results thoroughly and conduct a sensitivity analysis to assess the robustness and generalization capabilities of the learned autonomous driving strategies. The findings from this study contribute to the advancement of Gaussian process based Bayesian optimization to optimize the hyperparameters for autonomous driving in RL, providing valuable insights for the development of efficient and reliable autonomous driving systems.

\end{abstract}
\section{Introduction}

Autonomous Driving (AD), a transformative technology, holds the potential for revolutionizing transportation, enhancing safety, and improving efficiency. Deep Reinforcement Learning (DRL), prominent in various applications, including autonomous driving ~\cite{Spielberg2019TowardSP}, has shown promise in leveraging optimal neural network structures ~\cite{osti_1772592} to replace traditional rule-based controllers and enable model-free optimal control in complex environments ~\cite{li-sc17}. However, achieving optimal performance and stability during RL training necessitates careful tuning of hyperparameters such as discount factor, learning rate, batch size, network architectures, etc. These manually set parameters significantly influence the learning process and overall system performance. Hyperparameter Optimization (HPO) on RL poses several challenges due to the unique characteristics of RL algorithms and environments \cite{alshedivat2018continuous,Henderson_Islam_Bachman_Pineau_Precup_Meger_2018}. Some of the key challenges include: 
\begin{itemize}
    \item Sample Complexity: RL algorithms demand extensive interactions for effective policy learning. High sample complexity makes hyperparameter optimization computationally expensive, requiring evaluation across multiple training episodes
    \item Exploration-Exploitation Trade-off: Finding the right balance between exploration (trying out different actions to discover optimal strategies) and exploitation (taking actions that are known to yield high rewards) is crucial for effective learning, as overly conservative or aggressive settings hinder learning progress.
    \item High-Dimensional Search Space:RL involves a complex, high-dimensional search space with numerous hyperparameters. Increasing dimensionality challenges the search for optimal hyperparameters, making it difficult to explore various possible configurations.
    \item Non-Stationarity: RL environments exhibit non-stationarity, necessitating consideration of temporal aspects in hyperparameter optimization. Hyperparameters initially effective may become suboptimal as the environment evolves during the learning process.
    \item Sensitivity to Initial Conditions: RL algorithms are sensitive to initial conditions and hyperparameter changes. Small hyperparameter adjustments can lead to significantly different learning dynamics and performance, complicating the search for robust parameter sets.
    \item Interactions and Dependencies: Hyperparameters in RL are interdependent, making the search space complex. Adjusting one hyperparameter may impact the effectiveness of others, requiring careful consideration of interactions.
    \item Non-Convexity and Noisy Rewards: The RL optimization landscape is non-convex, meaning there can be multiple local optima that differ in performance. Moreover, rewards in RL can be noisy and have high variance, which makes it challenging to differentiate between the impact of hyperparameters and inherent randomness in the environment.
    \item Transferability: Hyperparameters effective in one RL task may not generalize, requiring task-specific fine-tuning. Achieving good performance across diverse RL problems adds complexity to hyperparameter optimization due to variations in task characteristics.
\end{itemize}

Balancing exploration and exploitation, leveraging prior knowledge, and utilizing computational resources effectively are key considerations in finding optimal hyperparameters for RL algorithms. This study highlights these issues as focal points for investigation. The paper aims to train an AD task through RL within a precisely controlled simulation environment using the Proximal Policy Optimization (PPO) algorithm. It seeks to optimize hyperparameters and investigate their significance, thereby contributing to a deeper understanding of effective parameter tuning for RL-based AD. The RL setting is treated as a blackbox function, where hyperparameters serve as inputs, and the objective is to maximize the rewards obtained by the autonomous vehicle. As direct optimization of the RL black-box function is computationally expensive, surrogate model optimization techniques are explored, providing a computationally efficient means to approximate black-box functions and effectively explore the hyperparameter search space \cite{10.5555/2986459.2986743}.  Our approach incorporates Efficient Global Optimization (EGO) \cite{jones1998efficient}, where probabilistic uncertainties are explained through Gaussian random fields. It is an iterative optimization method aiming to find the global minimum of a costly black-box function by sequentially evaluating new points based on a surrogate model. Given imbalanced parameters in our optimization problem, we emphasize the need to handle such cases by preprocessing the data using appropriate feature scaling techniques and efficiently exploring the high-dimensional search space with interdependencies between hyperparameters. Additionally, we address the non-convexity of the RL optimization landscape and noise in RL rewards, which pose challenges in accurately assessing hyperparameter impact. Moreover, sensitivity analysis is conducted on the hyperparameters to gain insights into their impact on the RL system's performance. Leveraging these techniques, our study contributes to designing robust, efficient, and reliable RL-based AD systems.

\section{Related Work}
Autonomous driving control has witnessed a growing interest in learning-based strategies, with RL showcasing notable advancements. RL excels in scenarios where predefined rules are ambiguous, leveraging environment information for decision-making. RL agents, capable of handling diverse situations through trial and error learning, offer advantages over manually designed policies \cite{10166271}.  Deep RL methods in autonomous driving enhance response time and exploit autopilot benefits \cite{wang2023efficient}. Despite this progress, RL agents can struggle in complex situations or demand extensive data. To overcome these challenges, RL algorithms can benefit from expert priors and motion skills, improving learning efficiency and driving performance \cite{wang2023efficient}. The combination of RL and rule-based constraints enhances autonomous vehicle safety \cite{10161418}. In this study, we use RL to learn driving behavior in a simulated environment via an external car controller with predefined rules. RL focuses on acquiring smooth driving control by learning the controller's behavior. Additional rule-based safety features are implemented for real-world tests. 

While machine learning models have demonstrated significant success across a wide range of applications \cite{article_fat,article_kand}, their performance heavily relies on the appropriate configuration of hyperparameters. As models become larger and more complex, the need for efficient and autonomous algorithms for tuning hyperparameters becomes increasingly critical to achieve optimal performance. Researchers explore diverse techniques for HPO, including black-box and multi-fidelity optimization methods \cite{ef762d1827e743799b725358e891b099}, aiming for automated tuning,  taking into account the intricacies and nuances of the underlying machine learning algorithms. HPO is a dynamic field; early methods like grid search and random search \cite{bergstra2012random} evolved into surrogate model-based Bayesian optimization \cite{ef762d1827e743799b725358e891b099} to address increasing complexity. Advances include improved surrogate models using heteroscedastic and evolutionary techniques \cite{cowenrivers2022hebo}, and attention to population-based and RL-based strategies. While existing literature extensively examines HPO methodologies, challenges, and future directions \cite{Feurer2019}, it lacks discussions specifically tailored to HPO in real-world RL contexts, with insufficient attention given to comprehensive sensitivity analysis. Prominent tools such as the HPO package  \cite{JMLR:v23:21-0888} employ Gaussian processes but do not explicitly cater to RL tasks, primarily not focusing on the initial Gaussian process fitting. Our approach allocates a significant computational budget for the initial fitting of the Gaussian process to account for the complete spread of the search space, thereby facilitating the optimizer's convergence to global optima. Despite the abundant literature on HPO, there's a noticeable scarcity of addressing hyperparameter optimization for RL algorithms at scale.

\section{Training Strategy Using RL}
\begin{figure}
\centerline{\includegraphics[width=0.8 \linewidth]{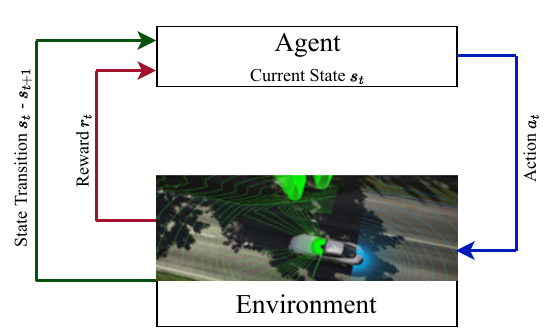}}
\caption{Framework of reinforcement learning} \label{rl}
\end{figure}

In RL \cite[p.~378]{aggarwal2018neural}, the agent (e.g., a real or simulated robot) takes actions to maximize the cumulative reward to solve a sequential decision task. RL seeks an optimal policy through trial and error interactions, addressing the agent-environment interface, reward evaluation function, and Markov property \cite{aggarwal2018neural}. The learning process involves mapping from states $s$ to actions $a$ to maximize rewards $r$. The agent receives sensory input as a state $s_t$ and, upon taking action $a_t$, obtains a reward $r_t$.  This loop stabilizes the algorithm. Trial and error aid exploration for better actions, while the agent's memory of successful actions ensures effective decision exploitation. Figure \ref{rl} depicts the RL agent interacting with its environment.

\subsection{Learning in Simulation}
\begin{figure} 
\centerline{\includegraphics[width= 250pt]{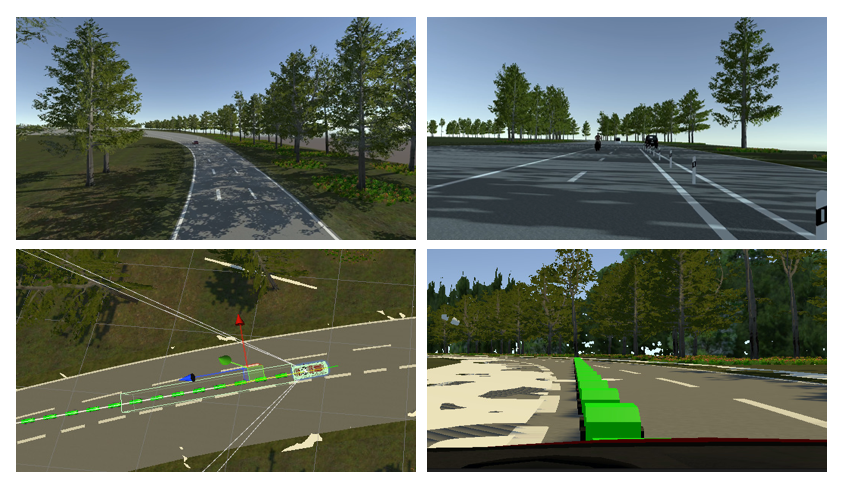}}
\caption{Unity3D simulator training environment. Top left: Drone view of road with agent. Top right: Camera recordings. Bottom: Driving behavior based on trajectory points.}
 \label{snippet}
\end{figure}

As it is tedious and expensive to train an autonomous vehicle directly in the real world, we employ the power of the Unity3D simulator \cite{Unity3d} to replicate the real-world scenario. The goal of the experiment is to train autonomous agents in simulation and test their behavior at a large three-lane oval proving ground (test track) available at our facility. Therefore, we aim to construct an exact replica of the real world, prioritizing the inclusion of minute features found in the actual environment. Multiple agents are trained on a simulated oval proving ground, aiming to navigate smoothly, minimizing jerks and abrupt movements and avoiding collisions.  Despite the seemingly simple task, the control system is sensitive, requiring the algorithm to learn the controller behaviour. Figure \ref{snippet} depicts the simulated training environment. Stationary obstacles such as cones and barricades, as well as movable obstacles like cars and bikes, are randomly spawned at various locations within the simulation. These obstacles remain in place until encountered by the agent. The agent's task is collision-free navigation around the proving ground using a camera-only approach, where the front camera's semantic segmentation serves as input to the neural network. The vehicle adopts a trajectory-based driving approach, with the external car controller executing the trajectories' driving behavior. The car controller oversees 23 trajectory points, managing velocity, acceleration, and heading. It manipulates gas, brake, and steering through a black box function detailed in section \ref{sec:inp}. Due to the controller's restrictive value acceptance, its behavior is learned through DRL, outlined in section \ref{sec:rew} via the reward function.

\subsection{RL Algorithm and Network Architecture}
\begin{figure} 
\includegraphics[width=\linewidth]{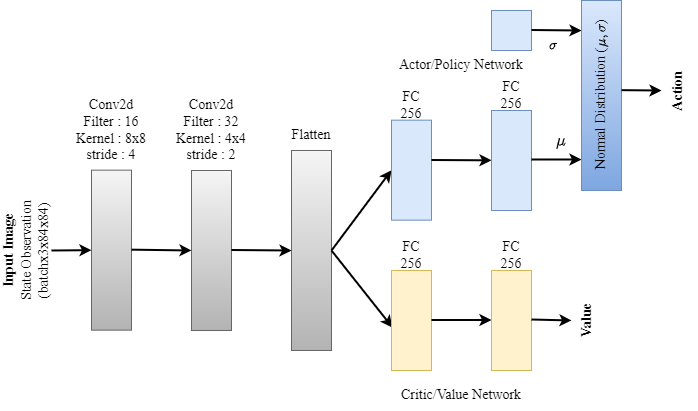}
\caption{PPO neural network architecture}
\label{PPOarch}
\end{figure}

In this research, we employ an Actor-Critic model with the PPO algorithm \cite{ppo}. In on-policy methods like PPO, the agent learns from data collected during interactions with the environment under its current policy. Using a shared convolutional component in a multi-head network, one head functions as the actor model, outputting a policy as a normal distribution by learning the mean and standard deviation of the actions. The other head serves as the critic or value function. The convolutional layer takes in semantic segmentation images of size $84 \times 84 \times 3$ as an input. Figure \ref{PPOarch} shows the neural network architecture of the PPO algorithm used. The network consists of 2 convolutional layers (conv2d) which are flattened and sent to the actor and critic blocks. Both the actor and critic networks employ fully connected layers (FC) with 256 units across two layers. The filters, kernels, stride and FC units are carefully selected after several trials and errors. The network is optimized using the Adam optimizer.

\subsubsection{Selection of Hyperparameters}
 As per \cite{ppo}, there are few commonly used hyperparameters contributing to the algorithm performance. Hence we focus on refining six key hyperparameters: batch size, time horizon, discount factor, learning rate, PPO epoch, and entropy beta (entropy coefficient) \cite{sutton2018reinforcement,ppo}. Learning rate and entropy beta undergo gradual discounts as the learning process advances. The initialization ranges for these hyperparameters are detailed in Table \ref{tab:hyp}. The remaining hyperparameters either maintain a constant value or undergo discounting. We set the initial PPO clipping parameter $\epsilon$ \cite{sutton2018reinforcement} value to $0.2$, reducing it to $0.1$ during RL training. This reduction aims to prevent the algorithm from taking larger steps as training progresses, thereby mitigating the risk of PPO deviating from the policy. The critic discount, which aims to equalize the significance of both the policy and value functions' error terms, is fixed at a constant value of $0.5$. 

 \begin{table}
\centering
\caption{Hyperparameter and its selected ranges for LHS}
\begin{tabular}{ll}
\textbf{Hyperparameter} & \textbf{Ranges}       \\
Batch Size     & 512 - 2560      \\
Time Horizon   & 64 - 600        \\
Discount       & 0.90 – 0.99     \\
Learning rate      & 0.00001 – 0.001 \\
PPO Epochs     & 3 - 10          \\
Beta           & 0.0001 – 0.01  
\end{tabular}
\label{tab:hyp}
\end{table}

\paragraph{Description of the chosen Hyperparameters}
\begin{itemize}
    \item Batch Size: The batch size determines the number of transitions or samples used in each policy update during training. It represents the amount of data the agent collects from the environment, consisting of observed states, actions, resulting next states, obtained rewards, and additional information. A larger batch size provides more accurate gradient estimates but requires more computational resources. Conversely, a smaller batch size reduces computational burden but may result in noisier gradient estimates. The data is typically sampled from a replay buffer, which is refilled after each PPO update since it is an on-policy method.
    \item Time Horizon: The time horizon corresponds to the number of time steps or interactions the agent considers when collecting trajectories during training. In each iteration of the PPO algorithm, the agent generates a set of trajectories by interacting with the environment. A trajectory comprises a sequence of states, actions, rewards, and possibly other information. The time horizon determines how far into the future the agent looks when generating these trajectories. The choice of the time horizon affects the balance between exploration and exploitation. A shorter time horizon focuses on immediate decision-making and facilitates quicker adaptation to environmental changes. Conversely, a longer time horizon allows the agent to consider more future states and rewards, capturing complex dependencies and enabling better long-term planning.
    \item Discount Factor: The discount factor, denoted as $\gamma$, influences the weight assigned to future rewards in the reinforcement learning algorithm. Ranging between 0 and 1, the discount factor determines how much the agent values immediate rewards compared to future rewards. A value of 0 means that only immediate rewards are considered, while a value of 1 treats future rewards equally to immediate rewards. When computing the cumulative return or total reward for a trajectory or episode, the discount factor is applied to discount the value of future rewards. This encourages the agent to consider the long-term consequences of its actions and promotes the optimization of policies that maximize cumulative rewards over time.
    \item Learning Rate: The learning rate controls the magnitude of the parameter updates during the optimization process. It determines the step size at which the model parameters are adjusted based on the gradients computed during training. The learning rate is a scalar value that influences the speed and stability of convergence. A higher learning rate can lead to faster convergence, but it may also cause overshooting or instability. Conversely, a lower learning rate ensures more cautious updates but may require a larger number of iterations to reach convergence.
    \item PPO Epoch: Epoch updates refer to the process of performing multiple iterations or updates of the policy network within a single epoch during the training phase. An epoch in PPO typically consists of several epoch updates. During each epoch update, the policy parameters are adjusted to improve the performance of the policy. The updates aim to maximize the objective function, which is often the expected cumulative reward or a surrogate objective that approximates it.
    \item Entropy Beta: The entropy beta parameter $\beta$ influences the exploration-exploitation trade-off in policy optimization. Entropy measures the uncertainty or randomness in an agent's policy. A higher entropy indicates greater uncertainty and decreases as the policy becomes more deterministic. In PPO, the entropy beta parameter controls the strength of the entropy regularization term in the objective function. This term encourages exploration by favoring policies with higher entropy, indicating greater randomness or uncertainty. By including this term, PPO promotes exploration, preventing premature convergence to suboptimal local optima and supporting the discovery of more optimal policies.
\end{itemize}

\subsection{Input and Output Specification}
\label{sec:inp}
The driving relies solely on a camera setup, using an $84 \times 84 \times 3$ semantic segmentation image from the front camera. Selection of a smaller image dimension is intended to reduce computational latency in the vehicle. This choice is also influenced by the ability to achieve satisfactory outcomes using smaller images within the given budget. Our approach leverages the ground truth segmentation images available within the simulation environment instead of predicting them. During inference/deployment, the car camera employs a trained perception network to convert the captured images into their corresponding semantic masks, which are then utilized as input observations for the network. As perception networks inherently contain a component of error, we deliberately introduced minor inaccuracies into our ground truth semantic segmentation images, incorporating approximately 5\% false segmentations. These inaccuracies were observed from the perception network and, based on this, it focuses on adding inaccuracies mainly at the edges of the semantic masks (primarily on the road surface).

The network generates trajectory points used by the external car controller to guide the vehicle. To maintain fidelity between simulation and reality, an accurate replica of the actual car controller is employed in the simulation during training. The black-box nature of the car controller requires the RL algorithm to learn its operation through rewards. The car controller component of the autonomous driving system receives a set of 23 trajectory points as input, with each point consisting of 10 values encompassing positions, angles, heading, curvature, lateral displacement, and other relevant parameters, resulting in a total of 230 values. Generating and learning 230 trajectory points directly poses challenges, requiring a large network size and extensive training. To address this, we simplify the output to 6 trajectory points, each conveying essential x-axis position and velocity. Consequently, the network outputs a total of 12 values responsible for determining the placement of these 6 trajectory points. We reconstruct the full set of 23 points using Catmull-Rom spline interpolation \cite{catmull1974class}. Subsequently, further post-processing steps are conducted to calculate additional values such as heading, curvature, and other relevant parameters, resulting in a total of 230 values corresponding to the complete set of 23 trajectory points.  These are inputs to the car controller, guiding the autonomous vehicle's behavior. Through training, the network learns to position these points for smooth driving, guided by a complex reward function detailed in the next section.

\subsection{Rewards}
\label{sec:rew}
In RL algorithms, rewards act as the objective function, guiding the agent's decision-making by evaluating the desirability of different states and actions. This enables the RL algorithm to discern and reinforce behaviors leading to higher rewards while discouraging actions resulting in lower rewards \cite{aggarwal2018neural}. The reward function utilized in this study is a comprehensive function that incorporates multiple factors to address two crucial aspects: the correct behavior of the external car controller and the driving behavior itself. The network must accurately position trajectory points to align with the external car controller's expectations, considering the controller's restrictions on value acceptance. Negative rewards are assigned if the controller rejects a trajectory, discouraging such actions.  Positive rewards are granted for successfully placing these points in the correct positions, aiming to steer the network towards accurate predictions. 
These rewards are crucial, especially in early training, motivating the network to produce meaningful trajectory points. The second element of the reward function primarily addresses driving behavior, penalizing undesirable actions like collisions and driving on non-drivable surfaces, while rewarding adherence to the correct path. Although these rewards are effective for guiding the vehicle, additional subtle aspects must be considered to achieve smooth driving without jerks. This reward component is designed based on \cite{kleen2012beherrschbarkeit}, which introduces a disturbance scalar and rates the driving behavior accordingly. Training the network with crafted reward functions successfully yields meaningful trajectories. However, achieving success requires meticulous tuning of hyperparameters impacting the model's performance, which will be explored in the next section.

\section{Model-Based Hyperparameter Optimization}
\begin{figure}
\centerline{\includegraphics[width= 250pt]{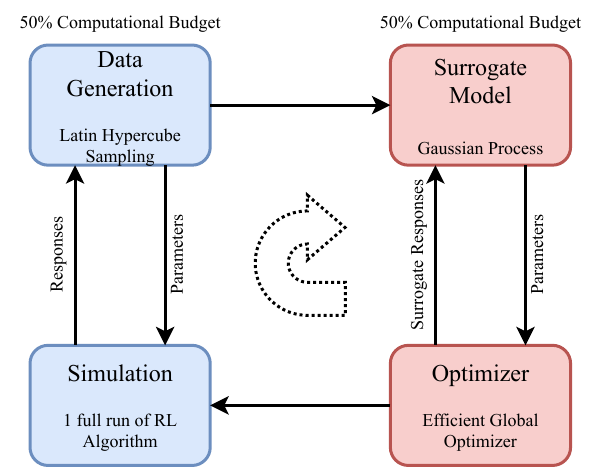}}
\caption{Optimizing hyperparameters through iterative refinement: Latin Hypercube Sampling initiates the search, while Efficient Global Optimizer maximizes cumulative rewards.} \label{opt_loop}
\end{figure}

Model-based hyperparameter optimization enhances machine learning model performance by constructing a surrogate model to approximate the hyperparameter-performance relationship \cite{Feurer2019}. It iteratively refines the search space to find optimal hyperparameters through an initialization and optimization process. In the initialization phase, a sampling method like Latin Hypercube Sampling (LHS) samples initial hyperparameter sets, and is evaluated for corresponding responses. A surrogate model is built from these observations. In the optimization phase, acquisition functions guide the surrogate model to identify the hyperparameter configuration likely to yield the best performance. The chosen configuration is evaluated, and optimized with an optimizer like Efficient Global Optimizer (EGO) and the iterative process continues until a satisfactory solution is found or a termination criterion is met. The optimization aims to maximize cumulative rewards in the RL algorithm by balancing exploration and exploitation for better performance. Figure \ref{opt_loop} illustrates the complete optimization process. 25-50\% of the computational budget is allocated to data generation, with the remaining budget dedicated to the optimization phase.

\subsection{Search Space Exploration - Latin Hypercube Sampling}
\label{sec:latin}
Latin Hypercube Sampling (LHS) \cite{mckay1979comparison} is a statistical technique for generating representative samples in a multidimensional parameter space. Unlike traditional random sampling, where parameters are randomly chosen within their ranges, LHS employs a stratified sampling scheme where the sampled points are organized in a grid-like structure that ensures superior coverage of the parameter space. Each parameter range is divided into equally sized intervals or bins, and a single value is sampled from each bin. This approach reduces bias, facilitates efficient exploration, enhances representativeness, and promotes diversity across the parameter space, especially in high-dimensional scenarios. The hyperparameters and their ranges are detailed in Table \ref{tab:hyp}.

\subsection{Gaussian Process}

In statistics, Kriging \cite{krige1951statistical} is an interpolation method based on Gaussian processes (GP) with prior covariances. The covariance of the Gaussian process represents uncertainties, providing not only the predicted mean of the objective function but also an estimate of the associated uncertainty. In HPO, GP interpolation probabilistically models the performance landscape, aiding in selecting which hyperparameter configurations to explore and exploit \cite{rasmussen2006gaussian,cowenrivers2022hebo}.
\par
These models are characterized by a mean function $\mu(x)$ and a covariance function (kernel) $k(x, x)$, which encodes the smoothness assumptions on $f(\cdot)$. Given a finite set of input $x_{1:n_i}$, the outputs follow a joint Gaussian distribution:
\begin{equation}
    f(x_{1:n_i})|\theta \sim N(\mu (x_{1:n_i}), K_{\theta}(x_{1:n_i},x_{1:n_i})),
\end{equation}
where $[\mu (x_{1:n_i})]_k = \mu(x_k)$ denotes the mean vector and $K_{\theta}(x_{1:n_i},x_{1:n_i})) \in R^{n_i \times n_i}$ the covriance matrix or kernel. Here, $k_{\theta}(.,.)$ depicts a parameterised kernel with unknown hyperparameters $\theta$ corresponding to lengthscales. We adopt a zero-mean prior notation, following conventions from \cite{rasmussen2006gaussian}. When choosing a GP kernel, options include the squared exponential, Gaussian, Matérn kernels, etc., each representing different prior assumptions about the latent function \cite{cowenrivers2022hebo}.
\par
Given the data points $\mathbf{X}_i$, assuming a Gaussian distributed observation with noise $y_i = f(x_i) + \epsilon$, where the noise is given by $\epsilon \sim \mathcal{N}(0,\sigma^2)$, the joint distribution over the data with an evaluation input $x$ can be written as \cite{rasmussen2006gaussian}:
\begin{equation}
    \left[\begin{array}{c}
        {y}_{1: n_i} \\
        f({x})
    \end{array}\right] \mid {\theta} \sim \mathcal{N}\left(\left[\begin{array}{c}
        \mu\left({x}_{1: n_i}\right) \\
        \mu({x})
    \end{array}\right], \right. \\
    \quad \left.\left[\begin{array}{cc}
        {K}_{{\theta}}^{(i)}  + \sigma_{{noise }}^2 I & k_{\theta}^{(i)}({x}) \\
        {k}_{{\theta}}^{(i), \mathbf{T}}({x}) & k_{{\theta}}({x}, {x})
    \end{array}\right]\right),
\end{equation}
where ${K}_{{\theta}}^{(i)}={K}_\theta\left({x}_{1: n_i}, {x}_{1: n_i}\right)$ and ${k}_\theta^{(i)}({x})={k}_\theta\left({x}_{1: n_i}, {x}\right)$. In the setting of $q$ arbitrary evaluation points, we get ${f}\left({x}_{1: q}^{\star}\right) \mid \mathbf{X}_i, {\theta} \sim \mathcal{N}\left({M}_i\left({x}_{1: q}^{\star} ; {\theta}\right), {\Sigma}_i\left({x}_{1: q}^{\star} ; {\theta}\right)\right)$ with:

\begin{equation}
\begin{aligned}
{M}_i\left({x}_{1: q}^{\star} ; {\theta}\right) = {K}_{{\theta}}^{(i)}\left({x}_{1: q}^{\star}, {x}_{1: n_i}\right)\left({K}_{{\theta}}^{(i)}+\sigma_{{noise }}^2 {I}\right)^{-1} \\ \left({y}_{1: n_i}-\mu\left({x}_{1: n_i}\right)\right)+\mu\left({x}_{1: q}^{\star}\right) 
\end{aligned}
\end{equation}
\begin{equation}
\begin{aligned}
{\Sigma}_i\left({x}_{1: q}^{\star} ; {\theta}\right)={K}_{{\theta}}^{(i)}\left({x}_{1: q}^{\star}, {x}_{1: q}^{\star}\right)-{K}_{{\theta}}^{(i)}\left({x}_{1: q}^{\star}, {x}_{1: n_i}\right) \\ \left({K}_{{\theta}}^{(i)}+\sigma_{{noise }}^2 {I}\right)^{-1} {K}_{{\theta}}^{\mathrm{T},(i)}\left({x}_{1: q}^{\star}, {x}_{1: n_i}\right) .
\end{aligned}
\end{equation}
The final step in the GP pipeline is to determine the unknown hyperparameters $\theta$ given a set of observation $\mathbf{X}_i$. In standard GPs, the hyperparamter $\theta$ is computed by minimising the negative log marginal likelihood (NLML) leading to the following optimisation problem \cite{rasmussen2006gaussian}:
\begin{equation}
\begin{aligned}
\min _{{\theta}, \sigma_{{noise }}} \mathcal{J}\left({\theta}, \sigma_{\text {noise }}\right)=\frac{1}{2} \log \operatorname{det}\left({\left({K}_{{\theta}}^{(i)}+\sigma_{{noise }}^2 {I}\right)}_{{\theta}}^{(i)}\right)+ \\ \frac{1}{2}\left({y}_{1: n_i}-\mu\left({x}_{1: n_i}\right)\right)^{{\top}} {\left({K}_{{\theta}}^{(i)}+\sigma_{{noise }}^2 {I}\right)}_{{\theta}}^{(i),-1} \\ \left({y}_{1: n_i}-\mu\left({x}_{1: n_i}\right)\right)+ \frac{n_i}{2} \log 2 \pi
\end{aligned}
\end{equation}

The objective in the above equation represents a non convex optimization problem making it susceptible to local minima. Many optimization solvers from $1^{\textbf{st}}$ order to $2^{\textbf{nd}}$ order and non-derivative based optimization methods have been studied in the literature to solve the optimization landscape.

\subsection{EGO Optimization}
\cite{jones1998efficient} combined Gaussian processes with the Expected Improvement (EI) function for derivative-free optimization, creating the Efficient Global Optimization (EGO) algorithm. This paper describes EGO as outlined by \cite{jones1998efficient}.
\par
Let $F$ be an expensive black-box function to be minimized. We sample $F$ at the different locations $x_i$ yielding the responses $y_i$. We build a Kriging model with a mean function $\mu$ and a variance function $\sigma^2$. EI can be expressed as:
$
E[I(x)] = E[\max(f_{\min}- Y, 0)]
$
where $Y$ is the best known objective function so far following the distribution $\mathcal{N}(\mu(x), \sigma^2(x))$. EI is computed by taking the expectation over the predicted distribution of the GP. This involves integrating the improvement function over the predicted distribution, often approximated using Monte Carlo sampling or other analytical methods. Evaluation of the equation using an error function can be found in \cite{forrester2008engineering}.
Next, we determine our next sampling point as:
$
x_{n+1} = \arg\max_{x} E[I(x)]
$
We then test the response $y_{n+1}$ of our black-box function $F$ at $x_{n+1}$, rebuild the model taking into account the new information gained, and research the point of maximum expected improvement again. Algorithm 1 provides a concise overview of the EGO process.

\begin{algorithm}
\SetAlgoLined
\SetKwInOut{Input}{Input}
\SetKwInOut{Output}{Output}

\Input{Function $F$, number of iterations $N$, initital LHS sampling points}
\Output{Best known solution after $N$ iterations}

\For{$i = 1$ \KwTo $N$}{
Build surrogate model$(X,Y)$ based on sample vectors $x_i$ and $y_i$;\\
Compute $f_{min} = \min(Y)$;\\
Compute EI for all points $x$;\\
Choose $x_{i+1}$ that maximizes EI;\\
Probe the function at most promising point $x_{i+1}$ : $y_{i+1} = F(x_{i+1}) $;\\
Update $x_i$ and $y_i$ with the new information;\\
Increment $i$;
}
\Return{Best known solution after $N$ iterations}
\caption{EGO with EI Acquisition function}
\end{algorithm}

\par
EI measures potential performance improvement relative to the current best solution by assessing novel hyperparameter sets. The parallel version, $q\text{EI}$ \cite{ginsbourger2010kriging}, proposes $q$ sampling points simultaneously, accelerating optimization by evaluating multiple sets concurrently. This enhances exploration, handles non-Gaussian distributions, improves scalability for high-dimensional problems, and adds robustness to noisy observations \cite{wu2016parallel}.

\section{Experiment}\label{heads}

\subsection{Setting Up the Black-box Function}

The black-box function can be written as $f(x_1, x_2, ....x_n) = Y,$ where the response variable $Y$ signifies the maximum cumulative reward per episode. Each run of this blackbox training signifies a complete run of the PPO algorithm with a certain set of hyperparameters. To expedite the process, we use 5 agents to collect state observations, actions, and rewards concurrently. An early stopping strategy halts training if rewards don't improve after about 50 iterations. Otherwise, the algorithm continues training for 300 iterations. Due to uncertainty about maximum achievable rewards, training isn't prematurely stopped even if it appears converged. For PPO training, as data batches are extracted from the buffer for PPO  update, the buffer size is configured to be 20 times the batch size. After each PPO update, the learning rate and entropy beta decrease more for higher values within their ranges. This ensures that significant adjustments are made for larger ranges, which helps improve the training process.

\subsection{Black-box Optimization}

\subsubsection{Training infrastructure}
The algorithm is trained in a cloud environment with 4 Nvidia V100 32GB GPUs. During data generation, 8 RL jobs are run in parallel across 4 GPUs. In the EGO phase with $q\text{EI}$, as we select the value of $q=4$, 4 RL jobs are run in parallel across the GPUs.

After setting up the black-box function, initial observations are gathered by repeatedly running the black-box function with various sets of hyperparameters. The hyperparameter sets ($200 \times 6$) are selected through LHS, with a larger budget of 50\% allocated for generating the initial data due to RL's high unpredictability. Parallelly executing the black-box function to generate 200 responses, the surrogate model is fitted using Gaussian process interpolation with a Gaussian kernel $\exp (-\sum_{j=1}^{k} \theta_j {|x_i - x_j|}^2)$ and nugget effect (adds a small non-zero variance term which accounts for noise or measurement errors in the training data) as the covariance function \cite{forrester2008engineering}.  Here, the squared term ensures a smooth correlation with a continuous gradient. $\theta_j$ is a learnable hyperparameter which serves as a width parameter determining the reach of influence for a sample point.

In Table \ref{tab:hyp}, hyperparameters exhibit varying ranges. To enhance GP fitting, scaling is applied, utilizing the logit function for the discount rate and the common logarithmic function $\log_{10}$ for learning rate and PPO Beta. Scaling ensures balanced representation, improving GP model effectiveness.  Following a satisfactory fit, parallel EGO (4 runs) employs the $q\text{EI}$ acquisition function to maximize the response variable, identifying the most promising hyperparameter set. This stage utilizes the remaining computational budget, executing 200 training runs during optimization. We experimented with various optimization runs using different budgets for initialization and optimization phases. Starting EGO optimization with fewer initial datasets took significantly longer for the optimizer to find better hyperparameters, and the expected performance wasn't achieved. Based on these trials, we chose to run 400 total training sessions, allocating 50\% of the budget to initialization and the EGO phase.
\subsection{Analysis}
Following black-box optimization, we compare the maximum cumulative rewards from initial observations to those obtained through EGO optimization, evaluating overall improvement. A time study examines the duration of the optimization process. Additionally, sensitivity analysis using ANOVA \cite{ANOVA} explores how hyperparameters influence our function's performance, assessing the significance of parameters and their interactions.

\section{Results}
\subsection{Performance Analysis}

\begin{figure}
\begin{subfigure}[h]{0.5\linewidth}
\includegraphics[width=\linewidth]{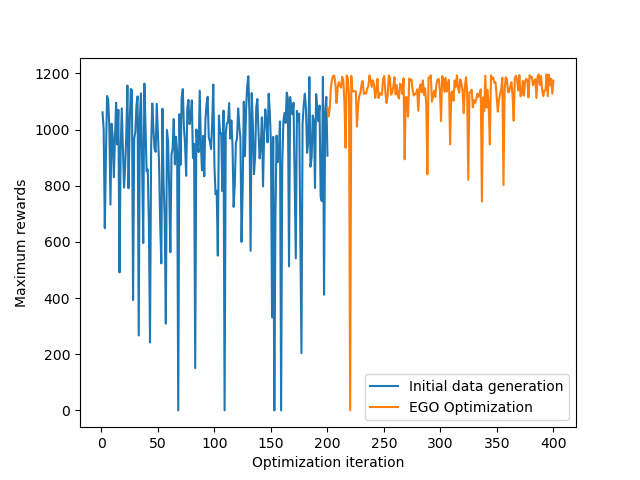}
\caption{}
\end{subfigure}
\begin{subfigure}[h]{0.5\linewidth}
\includegraphics[width=\linewidth]{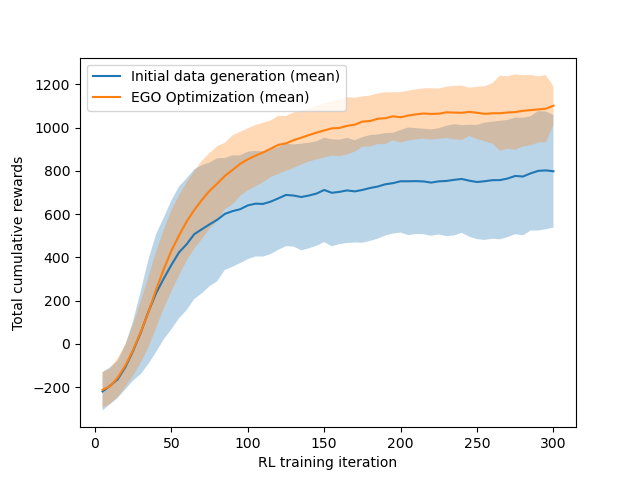}
\caption{}
\end{subfigure}
\caption{Reward Convergence plots. (a)Cumulative rewards achieved during the optimization process.  In the EGO phase, rewards are refined iteratively by tuning hyperparameters based on knowledge gained from previously evaluated data. (b) Reward Convergence for all RL iterations. Here, the average progression of rewards during RL training is plotted for both the initial data generation phase and the EGO phase. }
\label{rews}
\end{figure}

Firstly, the performance analysis assesses the optimization method's effectiveness by comparing the maximum cumulative rewards from initial observations with those from EGO phase. Figure \ref{rews}(a) displays the optimization history, depicting the progression of maximum rewards during iterative interactions. The blue line represents rewards from the data generation phase, while the orange line represents rewards from the optimization phase.  This visual representation allows for a clear understanding of the improvements achieved over the course of the optimization process.  Figure \ref{rews}(b) illustrates the convergence of rewards during RL training throughout the HPO iterations. Plotted every 5th iteration of the black box function (run for 300 RL iterations), the graph depicts mean rewards (blue line for data generation, orange for optimization). The faded blue shade indicates variability around mean rewards in the initial phase, while the faded orange shade represents standard deviation in the EGO phase. Mean and standard deviation are computed across every training run for each RL iteration (200 runs for initialization and 200 runs for optimisation). The graph demonstrates the optimization's learning process, highlighting its ability to generate optimal hyperparameters, resulting in a substantial improvement in cumulative rewards. In the EGO phase, rewards significantly increase compared to the initial data generation phase. Specifically at the $300^{th}$ iteration, the mean difference between the two phases is approximately 300 rewards. Moreover, the standard deviation of rewards in the EGO phase is considerably lower, indicating a more stable training behavior.

\begin{table}
\centering
\caption{Optimal hyperparameter achieved through HPO selected within their respective range.}
\begin{tabular}{lll}
\textbf{Hyperparameter } & \textbf{Range \hfill}  &  \textbf{    Optimal Value}       \\
Batch Size     & 512 - 2560     & $2029$  \\
Time Horizon   & 64 - 600       & $511$ \\
Discount       & 0.90 – 0.99     & $0.974$ \\
Learning rate      & 0.00001 – 0.001         & $9.3 \times 10^{-4}$ \\
PPO Epochs     & 3 - 10         & $3$\\
Beta           & 0.0001 – 0.01  & $0.01$
\end{tabular}
\label{tab:best_hyp}
\end{table}
The GP model built with collected initial observations, yielded a R-squared value of 0.48 through cross-validation. R-squared, also known as the coefficient of determination,  quantifies the goodness of fit of the model. After the optimization phase using the EGO, the final R-squared value increased to 0.69. The initial data generation phase through LHS is comparable to HPO through random search or grid search. Table \ref{tab:best_hyp} showcases the optimal hyperparameter set, achieving a maximum reward of 1193 at the $371^{st}$ iteration. The maximum cumulative reward achieved during the initialization phase was around 1140. The optimization led to a notable 4\% increase in the maximum reward compared to the initial data generation phase. The optimization history and improved R squared value demonstrate the efficiency of the process in enhancing the surrogate model's accuracy and maximizing rewards, yielding an optimal hyperparameter set. As expected, this improvement is reflected in the driving behavior. The autonomous agents showcased proficient navigation within the lane centre, relying solely on semantic segmentation information. Additionally, the network-generated trajectories demonstrated smoothness, avoiding abrupt changes and resulting in smooth acceleration and turns.

\subsection{Time Analysis}
The algorithm typically takes 6 to 16 hours, primarily influenced by the batch size. Considering 200 hyperparameter sets without early stopping, the initial data generation could take approximately 2200 hours. However, with 20\% of trainings stopping early and 8 parallel trainings with GPU utilization, this is reduced to 10 days. For EGO optimization with parallel $q\text{EI}$ acquisition, 4 parallel trainings lead to a total time of 20 days for 200 hyperparameter sets. In this case, early stopping is not as prevalent as EGO learns to generate hyperparameters that yield high cumulative rewards.

\subsection{Sensitivity Analysis}
\begin{figure}
\centerline{\includegraphics[width= 320pt, height = 200pt]{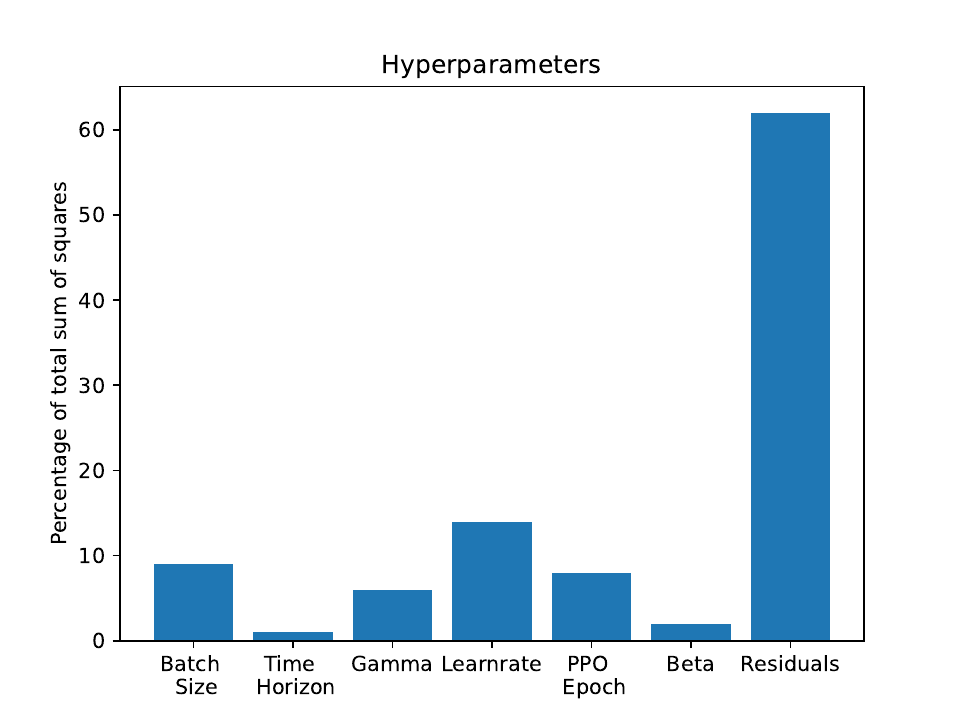}}
\caption{Percentage of total sum of squares} \label{sensitivity}
\end{figure}

This section summarizes the statistical performance of our hyperparameters, employing regression for model fitting and ANOVA for significance evaluation. A leave-one-out ablation study was also conducted, systematically excluding influential hyperparameters at each step to understand their sensitivity and interdependencies. This study helps identify the most impactful hyperparameters and assess their contributions to the model's predictive accuracy. Given the similarity in results, only the findings from the ANOVA are discussed. Table \ref{tab:sens} summarizes the hyperparameter influences, with ANOVA sums of squares (SS) indicating their respective impact. The F value column provides the result of the ANOVA F-test for each factor, with higher values signifying a more substantial effect on the dependent variable. Finally, Pr($>$F) column presents the p-values of the F test statistic associated with each factor to determine its significance. In the context of hypothesis testing in ANOVA, a p-value less than the chosen significance level (commonly 0.05) suggests that you have enough evidence to reject the null hypothesis. In summary, the null hypothesis states that all the hyperparameters seem to have a statistically significant impact on the dependent variable (reward). Compared to the other hyperparameters, the p-value of the time horizon is considerably higher making it less significant to others. Figure \ref{sensitivity} plots the percentage of the total sum of squares contributed by each hyperparameter.ANOVA and the ablation study suggest a descending order of influence: Learning rate (most influential) $>$ Batch $>$ PPO Epochs $>$ Gamma $>$ Beta $>$ Time Horizon (least influential). Time Horizon and Beta exhibit the lowest sensitivity, suggesting their variations have the least impact on performance.
\begin{table}
\centering
\caption{ANOVA of the linear model containing the sensitivity in the SS column}
\begin{tabular}{lrrrrr}
  \hline
 & Df & Sum Sq (SS) & Mean Sq & F value & Pr($>$F) \\ 
  \hline
Batch Size & 1 & 1469550.59 & 1469550.59 & 50.62 & 0.0000 \\ 
  Time Horizon & 1 & 129213.20 & 129213.20 & 4.45 & 0.0356 \\ 
  Gamma & 1 & 970001.90 & 970001.90 & 33.41 & 0.0000 \\ 
  Learning rate & 1 & 2430213.17 & 2430213.17 & 83.71 & 0.0000 \\ 
  PPO Epochs & 1 & 1393485.29 & 1393485.29 & 48.00 & 0.0000 \\ 
  PPO Beta & 1 & 319998.77 & 319998.77 & 11.02 & 0.0010 \\ 
  Residuals & 400 & 10480358.63 & 29031.46 &  &  \\ 
   \hline
\end{tabular}
 \label{tab:sens}
\end{table}

\section{Conclusion and Future work}
Our study optimized the PPO algorithm for AD through parallel EGO optimization, yielding a notable 4\% performance boost. The primary goal was to automate hyperparameter discovery in a cloud environment, eliminating the need for repeated RL algorithm runs through parallelization.  Sensitivity analyses provided valuable insights into parameter importance. This method shows potential for enhancing RL-based tasks by identifying optimal hyperparameter setups.
\par
In HPO, significant improvements are possible. A promising approach is multi-objective optimization to maximize rewards while minimizing computational resources for faster training. Our future research explores deep GPs and evolutionary algorithms as alternative optimization methods and further integrating network architecture search to optimize the overall performance. Furthermore, conducting a comprehensive benchmark comparison could provide valuable insights into performance metrics. To effectively evaluate driving performance, especially in complex scenarios with ambiguous rewards, a dedicated driving metric can be developed. Investigating the scalability and generalization of the proposed methods across various RL domains can offer valuable insights in understanding how well these methods perform in different contexts.

\begin{credits}
\subsubsection{\ackname} This work was supported by Volkswagen AG.

\subsubsection{\discintname}
The results, opinions and conclusions expressed in this publication are not necessarily those of Volkswagen Aktiengesellschaft.
\end{credits}

%
%

\bibliographystyle{splncs04}
\bibliography{refs.bib}

\begin{thebibliography}{10}
\providecommand{\url}[1]{\texttt{#1}}
\providecommand{\urlprefix}{URL }
\providecommand{\doi}[1]{https://doi.org/#1}

\bibitem{Unity3d}
Unity3d game engine. \url{https://unity.com/}

\bibitem{aggarwal2018neural}
Aggarwal, C.C.: Neural Networks and Deep Learning. A Textbook. Springer, Cham (2018). \doi{10.1007/978-3-319-94463-0}

\bibitem{alshedivat2018continuous}
Al-Shedivat, M., Bansal, T., Burda, Y., Sutskever, I., Mordatch, I., Abbeel, P.: Continuous adaptation via meta-learning in nonstationary and competitive environments (2018)

\bibitem{osti_1772592}
Balaprakash, P., Salim, M., Uram, T.D., Vishwanath, V., Wild, S.M.: Deephyper: Asynchronous hyperparameter search for deep neural networks  (1 2018)

\bibitem{10.5555/2986459.2986743}
Bergstra, J., Bardenet, R., Bengio, Y., K\'{e}gl, B.: Algorithms for hyper-parameter optimization. In: Proceedings of the 24th International Conference on Neural Information Processing Systems. p. 2546–2554. NIPS'11, Curran Associates Inc., Red Hook, NY, USA (2011)

\bibitem{bergstra2012random}
Bergstra, J., Bengio, Y.: Random search for hyper-parameter optimization. Journal of Machine Learning Research  \textbf{13},  281--305 (2012)

\bibitem{catmull1974class}
Catmull, E., Rom, R.: A class of local interpolating splines. In: Barnhill, R.E., Reisenfeld, R.F. (eds.) Computer Aided Geometric Design. pp. 317--326. Academic Press, New York (1974)

\bibitem{cowenrivers2022hebo}
Cowen-Rivers, A.I., Lyu, W., Tutunov, R., Wang, Z., Grosnit, A., Griffiths, R.R., Maraval, A.M., Jianye, H., Wang, J., Peters, J., Ammar, H.B.: Hebo pushing the limits of sample-efficient hyperparameter optimisation (2022)

\bibitem{article_fat}
Fatima, M., Pasha, M.: Survey of machine learning algorithms for disease diagnostic. Journal of Intelligent Learning Systems and Applications  \textbf{09},  1--16 (01 2017). \doi{10.4236/jilsa.2017.91001}

\bibitem{Feurer2019}
Feurer, M., Hutter, F.: Hyperparameter Optimization, pp. 3--33. Springer International Publishing, Cham (2019). \doi{10.1007/978-3-030-05318-5_1}

\bibitem{forrester2008engineering}
Forrester, A., Sobester, A., Keane, A.: Engineering Design via Surrogate Modelling: A Practical Guide. Wiley (2008)

\bibitem{ginsbourger2010kriging}
Ginsbourger, D., Le~Riche, R., Carraro, L.: Kriging is well-suited to parallelize optimization. In: Computational Intelligence in Expensive Optimization Problems, pp. 131--162. Springer, Berlin, Heidelberg (2010)

\bibitem{Henderson_Islam_Bachman_Pineau_Precup_Meger_2018}
Henderson, P., Islam, R., Bachman, P., Pineau, J., Precup, D., Meger, D.: Deep reinforcement learning that matters. Proceedings of the AAAI Conference on Artificial Intelligence  \textbf{32}(1) (Apr 2018). \doi{10.1609/aaai.v32i1.11694}

\bibitem{jones1998efficient}
Jones, D.R., Schonlau, M., Welch, W.J.: Efficient global optimization of expensive black-box functions. Journal of Global Optimization  \textbf{13}(4),  455--492 (1998)

\bibitem{ef762d1827e743799b725358e891b099}
Kandasamy, K., Dasarathy, G., Schneider, J., P{\'o}czos, B.: Multi-fidelity bayesian optimisation with continuous approximations. pp. 2861--2878. International Machine Learning Society (IMLS) (Jan 2017)

\bibitem{article_kand}
Kandasamy, K., Neiswanger, W., Schneider, J., Poczos, B., Xing, E.: Neural architecture search with bayesian optimisation and optimal transport  (02 2018)

\bibitem{kleen2012beherrschbarkeit}
Kleen, A., Vollrath, M.: Beherrschbarkeit von komplexen eingriffen in die fahrzeugführung. In: VDI Fahrzeug- und Verkehrstechnik (Hrsg.). Fahrerassistenz und Integrierte Sicherheit. 27. VDI Verlag GmbH, Düsseldorf (2012)

\bibitem{krige1951statistical}
Krige, D.G.: A statistical approach to some basic mine valuation problems on the witwatersrand. Journal of the Chemical, Metallurgical and Mining Society of South Africa  \textbf{52}(6),  119--139 (1952)

\bibitem{li-sc17}
Li, Y., Bel, O., Chang, K., Miller, E.L., Long, D.D.E.: {CAPES}: Unsupervised storage performance tuning using neural network-based deep reinforcement learning. In: Supercomputing '17 (Nov 2017)

\bibitem{JMLR:v23:21-0888}
Lindauer, M., Eggensperger, K., Feurer, M., Biedenkapp, A., Deng, D., Benjamins, C., Ruhkopf, T., Sass, R., Hutter, F.: Smac3: A versatile bayesian optimization package for hyperparameter optimization. Journal of Machine Learning Research  \textbf{23}(54), ~1--9 (2022)

\bibitem{mckay1979comparison}
McKay, M.D., Beckman, R.J., Conover, W.J.: A comparison of three methods for selecting values of input variables in the analysis of output from a computer code. Technometrics  \textbf{21}(2),  239--245 (May 1979). \doi{10.2307/1268522}

\bibitem{rasmussen2006gaussian}
Rasmussen, C.E., Williams, C.K.I.: Gaussian Processes for Machine Learning, Vol. 2. MIT Press, Cambridge, MA (2006)

\bibitem{ANOVA}
Sawyer, S.: Analysis of variance: The fundamental concepts. Journal of Manual \& Manipulative Therapy  \textbf{17},  27E--38E (04 2009)

\bibitem{ppo}
Schulman, J., Wolski, F., Dhariwal, P., Radford, A., Klimov, O.: Proximal policy optimization algorithms  (07 2017)

\bibitem{Spielberg2019TowardSP}
Spielberg, S.P., Tulsyan, A., Lawrence, N.P., Loewen, P.D., Gopaluni, R.B.: Toward self‐driving processes: A deep reinforcement learning approach to control. AIChE Journal  (2019)

\bibitem{10166271}
Sun, J., Fang, X., Zhang, Q.: Reinforcement learning driving strategy based on auxiliary task for multi-scenarios autonomous driving. In: 2023 IEEE 12th Data Driven Control and Learning Systems Conference (DDCLS). pp. 1337--1342 (2023)

\bibitem{sutton2018reinforcement}
Sutton, R.S., Barto, A.G.: Reinforcement Learning: An Introduction. The MIT Press, second edn. (2018)

\bibitem{10161418}
Udatha, S., Lyu, Y., Dolan, J.: Reinforcement learning with probabilistically safe control barrier functions for ramp merging. In: 2023 IEEE International Conference on Robotics and Automation (ICRA). pp. 5625--5630 (2023). \doi{10.1109/ICRA48891.2023.10161418}

\bibitem{wang2023efficient}
Wang, L., Liu, J., Shao, H., Wang, W., Chen, R., Liu, Y., Waslander, S.L.: Efficient reinforcement learning for autonomous driving with parameterized skills and priors (2023)

\bibitem{wu2016parallel}
Wu, J., Frazier, P.: The parallel knowledge gradient method for batch bayesian optimization  (2016)

\end{thebibliography}

\end{document}